# Applying NLP to iMessages: Understanding Topic Avoidance, Responsiveness, and Sentiment

**Alan Gerber, Sam Cooperman**
Emory College of Arts and Sciences, Emory University

## Abstract

What is your messaging data used for? While many users do not often think about the information companies can gather based off of their messaging platform of choice, it is nonetheless important to consider as society increasingly relies on short-form electronic communication. While most companies keep their data closely guarded, inaccessible to users or potential hackers, Apple has opened a door to their walled-garden ecosystem, providing iMessage users on Mac with one file storing all their messages and attached metadata. With knowledge of this locally stored file, the question now becomes: What can our data do for us? In the creation of our iMessage text message analyzer, we set out to answer five main research questions focusing on topic modeling, response times, reluctance scoring, and sentiment analysis. This paper uses our exploratory data to show how these questions can be answered using our analyzer and its potential in future studies on iMessage data.

## 1 Introduction

There has already been a fairly long and rich history of studying short-form messaging platforms to determine which platforms and services can be used as legitimate corpora for Latent Dirichlet Allocation (LDA). While many of these studies have focused on rapid Twitch chats, massive Discord server chatrooms, or early SMS messaging full of unique slang and abbreviations, there is not much to be found on iMessage data. This is mainly due to Apple's vertically integrated ecosystem locking most of their users' data behind heavy encryption. However, Mac users can easily access a file holding all of their iMessage data that can be cleaned and analyzed, labeled chat.db.

Some applications already exist that are able to process this data and break it down into simple metrics such as messages sent versus received. However, our project, [textmessageanalyzer.com](textmessageanalyzer.com), takes this a step further by utilizing Gensim topic modeling, reluctance score calculations, and VADER sentiment analysis to provide users with answer to the following questions:

1) What topics do you tend not to engage with?
2) What are your most commonly discussed topics?

3) How does your responsiveness change between one-to-one messaging and group messaging?
4) What topics tend to start your conversations?
5) What is the sentiment breakdown of your messages?

This all runs locally on the user's computer, keeping their data completely safe, while allowing them to see important insights in their online activity, such as how quickly they respond to messages both one on one and in a group chat or if they have generally become more positive or negative over time. Our application is also currently the only program that allows this data to be exported in a CSV file for custom analysis in Excel, R, and Python. Due to the nature of Gensim topic modeling often returning topics that do not translate cleanly into easy-to-digest topics, such as "Football" or "Movies", we have also included the ability for those with OpenAI API keys to feed this data directly to ChatGPT for more human-like topic analysis.

## 2 Literature Review

Before discussing our methodology and results, it is important to dissect what research has already been conducted in the fields of topic modeling, mobile data infrastructures, and the development of unique LDA (Latent Dirichlet Allocation) models.

### 2.1 Foundations of Messaging Data

Modern messaging platforms have evolved from basic SMS into encrypted, metadata-rich systems that define applications like iMessage. Tilson, Sorensen, and Lyytinen (2012) describe Apple's vertically integrated design philosophy as a "control paradox," one that restricts researchers' access to encrypted content while providing users with a standardized, locally stored chat.db file holding all of their messaging data. This structure simultaneously protects user privacy and dictates how digital communication research must be conducted, directly informing our own methodological approach.

### 2.2 Text Analysis and Topic Modeling

Early work on short-form text analysis established the foundation for modern messaging studies. Al Moubayed et al. (2016) demonstrated that probabilistic topic models combined with deep learning could extract consistent patterns from informal SMS data. Their findings expanded the viability of short, fragmented text as a legitimate corpus for unsupervised learning. Blei, Ng, and Jordan's introduction of LDA (2003) remains the central probabilistic framework for deriving latent thematic structures, while Zhao et al. (2011) proved that modified LDA techniques could effectively model brief, high-volume messages such as tweets. Parallel to content-based studies, Nematzadeh et al. (2019) revealed that Twitch users tend to disengage when the chat becomes overloaded with users and messages, mirroring our study into the effect of iMessage group chat size on responsiveness.

Rehurek and Sojka's Gensim (2010) and McCallum's MALLET (2002) remain the primary infrastructures for efficient,

large-scale topic modeling, balancing speed and precision. For our purposes, Gensim proved to be more practical as it specializes in short but numerous text segments compared to MALLET, which is built to handle longer document style corpora.

## 2.3 iMessage-Specific Insights

Research directly focused on iMessage further refined our methodological focus. Govan (2013) documents Apple's timestamp behavior and its unique 128-second server grouping, enabling temporal modeling with greater accuracy. Tiner and Anderson (2018) clarify iMessage's cryptographic design, distinguishing what remains encrypted compared to what metadata is locally accessible, and thus will be part of this project's collected data. Extending across multiple platforms, Ramos et al. (2023) demonstrate that response rhythms and time-of-day patterns consistently reveal interpretable behavioral trends, supporting our project's exploration of user reluctance and engagement.

Together, these studies establish the theoretical and technical basis for our work: that iMessage is uniquely positioned for meaningful, privacy-respectful analysis through local metadata, and that refined topic modeling can uncover behavioral insights even within fragmented, short-form digital communication.

## 3 Methodology

Our methodology includes three parts: data acquisition and preprocessing, topic modeling and sentiment analysis, and metrics calculations.

## 3.1 Data Acquisition and Preprocessing

For this project, we analyzed iMessage data from both authors of this paper, Alan and Sam, as well as an Emory University classmate who wished to stay anonymous. The data for each participant was contained within the iMessage chat.db file. The size of the file is determined by a multitude of factors such as how much iCloud data the user has access to, iMessage settings, or device age, with all three participants having file sizes between 0.8 and 1.0 gigabytes.

iMessage data comes in a very unclean format which requires thorough data processing. This meant stripping fragments of Apple metadata from the text, totaling 78 words such as "bplist", "tdate", or other terms which do not describe textual data. Then the chat history is converted into a dataframe containing chat group size, tapbacks such as likes and hearts converted as separate text messages, message direction (inbound/outbound), participant ID, timestamp, and content. This processed CSV of user text messages is free for export under the "Export Data" section of the deployable tool.

The full source code for the iMessage analyzer is publicly available in our GitHub repository (Gerber, 2025).

The text contents were converted into lowercase format, and punctuation removed. Lemmatization and English stopword removal were avoided in this analysis due to the short-form nature of text messages and our preference to preserve as much data as possible for our analysis.

### 3.2 Topic Modeling and Sentiment Analysis

We trained LDA topic modeling using Gensim on the aggregate text messages of each user (Rehurek and Sojka, 2010). Sentiment analysis was conducted on the text message data using VADER sentiment analysis (Hutto and Gilbert, 2014). VADER and its rule-based approach are great for text message data due to the fast application and ability to process emojis and punctuation without any preprocessing required.

### 3.3 Metric Calculations

Each received text message (only inbound) was given a reluctance score which is simply the number of minutes taken to respond to a text divided by 1440 (the number of minutes in a day), with a score cap of 1.0. This score was then multiplied by the topic probability for a topic given a certain text, and then divided by the topic probability. These values were all summed to produce an average reluctance metric for each topic.

$$\text{Average Reluctance} = \frac{\Sigma(topic\_prob \times reluctance\_score)}{\Sigma(topic\_prob)}$$

Where:

- *topic_prob* = how much that message belongs to the topic

- *reluctance_score* = reluctance value for that message (minutes to respond/1440 minutes)

The average reluctance scores were then multiplied by the log of total frequency of texts per topic (number of texts that had >0.3 probability of a certain topic) in order to avoid certain outlier topics with low text frequency from being ranked as most reluctant.

To calculate topic prevalence over time, the assigned topic probability of all text messages (inbound and outbound) within a certain time frame was averaged together.

Reply rates were calculated by marking all texts that had a subsequent reply within 1440 minutes OR a tapback reply OR a threaded response as having received a reply. Reply rates for different group size categories were calculated by finding what proportion of all messages for each group received a reply. Median response times were calculated by finding response times only for messages where there was a reply from the user.

To quantify which topics are most effective at initiating conversations, we computed a *starter_score* that combines the reply rate, speed, and frequency of messages for each topic.

$$\text{starter\_score} = (0.4 \ast reply\_rate + 0.3 \ast speed\_score + 0.3 \ast starter\_prob)$$

Where:

- $reply\_rate = \frac{\#messages\ within\ a\ topic\ with\ reply}{Total\ Messages\ within\ topic}$
- $speed\_score = 1 - \frac{average\ response\ time\ within\ topic}{1{,}440\ minutes}$
- *starter_prob* = proportion of messages within a topic that are conversation starters (first message in chat or at least 3 hours from previous)

The top 10 topics by *starter_score* are then graphed.

Total sentiment for user's text messages is computed using standard VADER thresholds with scores above 0.05 considered positive, below -0.05 considered negative, and between -0.05 and 0.05 as neutral. The sentiment for all messages within a given time span is then categorized and graphed.

## 4 Results and Figures

The following results reveal the ability of topic modeling, sentiment analysis, and our established metrics to discern user behavior in iMessage data. To better understand the data, Alan's iMessages contained 525,655 text messages starting in 2018, Sam's data contained 524,737 text messages starting in 2019, and our anonymous participant's data contained 422,801 messages starting in 2022. These are roughly comparable datasets, yet still varied enough to the point where text volumes may influence certain results

### 4.1 Reluctance

The most reluctant topics per individual varied greatly, but were ranked according to their respective Average Reluctance final score. To further understand these topics, outside of just the top 30 words for each, our tool allows for examining high reluctance messages that are categorized into each topic. With Alan, it was evident through looking at these examples that his high reluctance messages in topic 27 tended to be those containing time-pertinent information (Figure 1). For Sam, his most reluctant messages for topic 29 were related to fantasy football (Figure 2). For our anonymous participant, her most reluctant messages from topic 18 tended to be from school project group chats (Figure 3). Following the guidelines of recent studies, we chose to decipher these topics by analyzing these most related topic text messages, instead of just top words per topic, as this tends to yield superior results than relying solely on Gensim output (Gillings and Hardie, 2023).

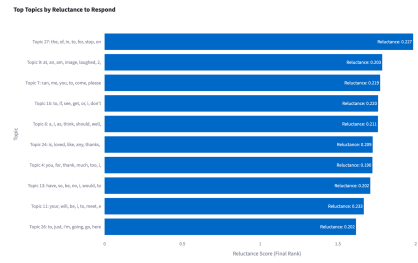

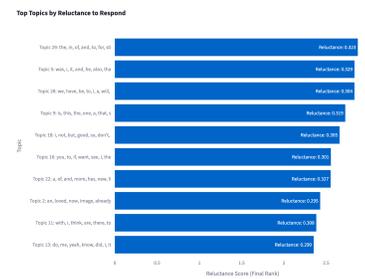

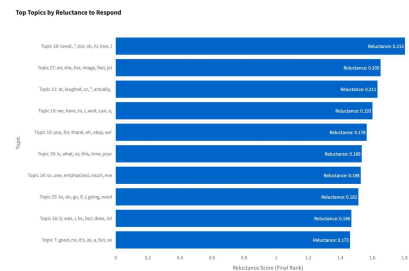

**Figures 1, 2, 3.** Most reluctant topics for Alan (Figure 1), Sam (Figure 2), and an anonymous Emory student (Figure 3)

### 4.2 Topic Prevalence

To better understand how topics of conversation may change over time for users, we examined the average topic prevalence for each topic over time. The general trend, barring certain exceptions, is for all discussed topics to remain relatively constant over time as seen in Figures 4-6. As can be seen especially in the results of Sam and our anonymous participant, out of 30 topics, none had exceeded a 0.1 topic prevalence at any point in their respective chat histories (Figures 5 and 6). In Alan's data within Figure 4, we see one topic break this rule, but this may largely be skewed by the exceptionally low iMessage volume of this period (February 2018 - June 2019), where average text volume was 15 per day, compared to his current 180 per day. This data shows how overall topic distributions tend to stay relatively constant over time. However, certain topics do tend to predominate over others, revealing usage patterns for iMessage. One example can again be seen in Figure 4 where topic 27 remains consistently higher than Alan's other message content. This topic contains administrative language such as "sale", "Saturday", "due" or "office" allowing us to see a pattern of iMessage used heavily for work tasks in this case. These results may help users identify what their primary iMessage use cases are, and how they change over time.

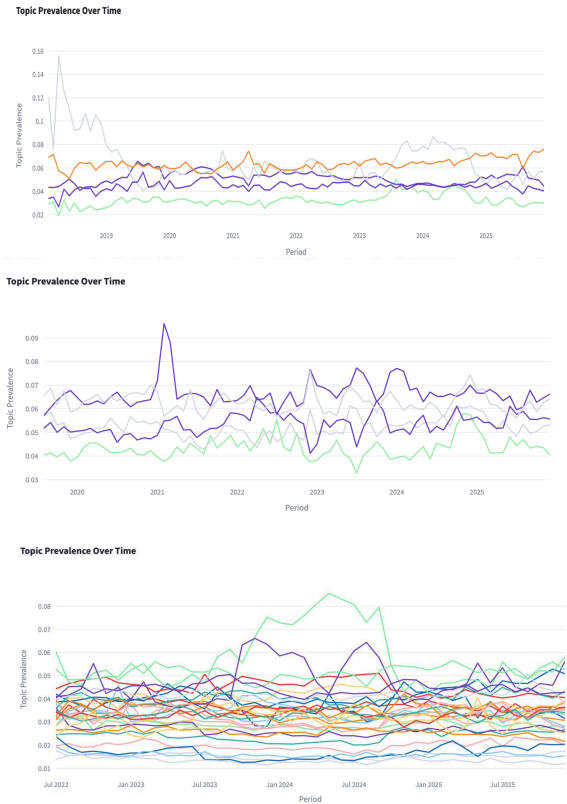

**Figures 4, 5, 6.** Top 5 most prevalent topics for Alan (Figure 4), Top 5 most prevalent topics for Sam (Figure 5), All topics for anonymous Emory student (Figure 6)

### 4.3 Group Chat Behavior

Previous research has shown in various cases how text responsiveness tends to decrease as group sizes increase (Fatkin and Lansdown). Our results further support these previous findings by showing how these results do, for the most part, manifest themselves in iMessage texts. Across all participants, Large group chats (9+ participants) had the lowest response rates, while one-to-one chats had the highest response rates (Figures 7-9). Additionally, median response times were lowest for one-to-one chats across all participants (Figures 7-9). However, there was still variance across participants for intermediate group sizes, and even greater variance in median response times. We can see with our anonymous participant that her median response times in group chats were a magnitude higher than either Alan or Sam's which may be due to different texting behaviors such as preferring to send one long text in large chats compared to multiple smaller texts (Figures 7-9). Additionally, having just one group chat that is hyperactive and/or highly time-sensitive is enough to throw off these aggregate trends. This is evidenced in Figure 9, where our anonymous participant, after further analysis, had a singular medium sized chat (5-8 participants) that was very active, yet very high reluctance, further skewing her response times upwards.

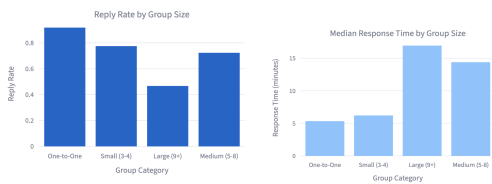

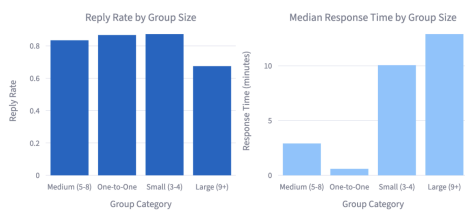

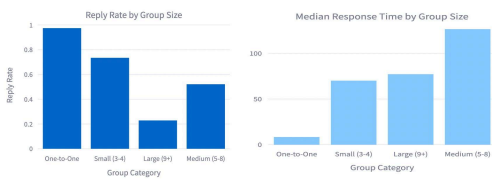

**Figures 7, 8, 9.** Reply rates and response time split by different group sizes for Alan (Figure 7), Sam (Figure 8), and an anonymous Emory student (Figure 9)

### 4.4 Conversation Starter Topics

A further point of analysis was to rank the LDA topics by which were most likely to initiate and continue conversations. Through our results, we see that there was not a significant difference between these "best" topics that really made one a better conversation starter than the other (Figures 10-12). However, a difference does still exist, and these can be seen with closer reading of individual topic top word results. In figure 10, Alan's topic 8 is the highest ranked, and this topic contains key words related to travel or activities. This suggests that travel may be a favored conversation starter of Alan's. Further analysis on each user's individual topic results and associated key-words may be helpful in identifying these favored points of discussion.

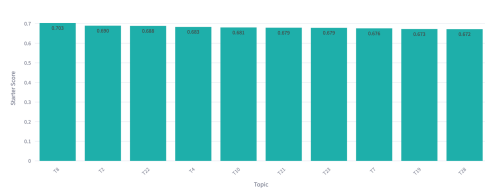

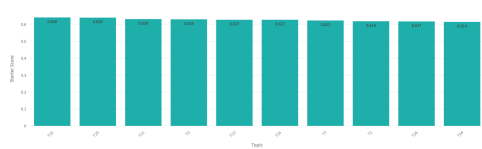

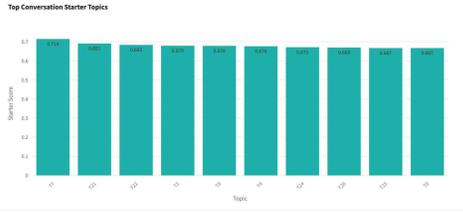

**Figures 10, 11, 12.** 10 top conversation starter topics for Alan (Figure 10), Sam (Figure 11), and an anonymous Emory student (Figure 12)

### 4.5 Sentiment Analysis

Text messages for each user were analyzed using VADER sentiment analysis and assigned a sentiment score from -1.00 to 1.00. These aggregate scores were then used to arrive at the insights in Figures 13-18. As we can see from all participants, neutral texts tended to predominate as seen in Figures 13-15. However, these similarities of sentiment do vary significantly between categorizations, as seen with the anonymous participant having more positive messages than neutral ones when looking at only outbound texts (Figure 15).

Additionally, we can see how the proportions of various sentiment messages change over time. Across all three participants, we can see a slightly upwards trend in positive messages over time, with the only exception to this being the 2018-2019 period for Alan due to exceptionally low text volume (Figures 16-18). This positive bias across all three participants, across all time frames, may reflect a larger trend that people tend to generally text more positively rather than negatively. These findings somewhat contradict previous studies which found social media platforms, such as twitter, tend to promote negative sentiment tweets (Schone et al., 2021).

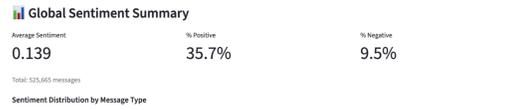
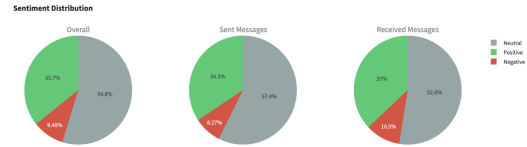
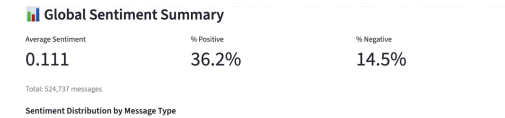
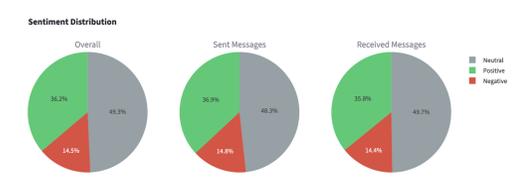
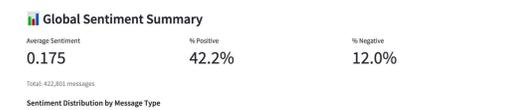
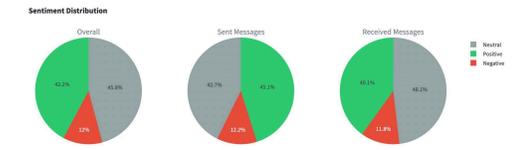

**Figures 13, 14, 15.** Text message sentiment summary statistics for Alan (Figure 13), Sam (Figure 14), and an anonymous Emory student (Figure 15)

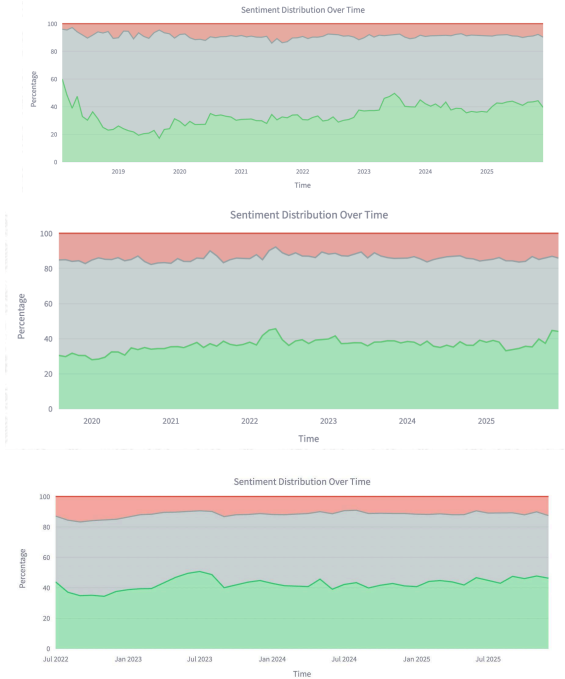

**Figures 16, 17, 18.** Text message sentiment change over time for Alan (Figure 16), Sam (Figure 17), and an anonymous Emory student (Figure 18)

## 5   Implementation and Future Directions

Our findings demonstrate that iMessage files can be analyzed using our iMessage-analyzer to arrive at novel findings regarding text behavior. Almost all of the functionalities and figures displayed may also be generated for individual chat conversations using this analyzer.

For reluctance, users can use the other functions of our analyzer to see which contacts they are most reluctant to text with. They can also take this analysis on an individual level to find over what periods of time they were most reluctant to respond to a particular contact and what their aggregate reluctance is for this contact. In future studies, these functionalities may be used to find common themes and topics that people tend to not respond to across multiple users.

Topic prevalence for individual conversations or group chats may also be tracked using in-built functions in our analyzer, to see which contacts a user may communicate to for different topics. In future works, topic prevalence changes, and more "defined" topics can be measured across multiple users to see how real-world events influence texting patterns.

There is additional functionality that lets users see which contacts they respond to more in groups, and even search individual contacts to find this information. This may help users know who to respond more often to in one-on-one group chats. Future studies may look into the analysis found in this paper across larger study cohorts to definitively state that there is a negative correlation between group chat size and user responsiveness among iMessage data.

In terms of topics most heavily associated with starting conversations, future functionality can be built to find how conversation starting varies by user. Future studies may look into how topics for conversation starters overlap in larger cohorts.

Further analysis for individual chat conversations using our analyzer is possible to help users identify who they should text more positively. Using our tool, the user can also see how the sentiment differs for inbound texts versus outbound texts within a certain chat, and how that changed over time. Future studies may look into tracking text-message sentiment to measure changes in user behavior such as potentially measuring or even diagnosing depression by examining text messages, compared to the current short-term studies tying depression to text behavior (Liu et al., 2022). Additionally, studies using large cohorts may reveal underlying trends in text behavior tending to be more positive than other forms of social media as our results may suggest.

Finally, every result that is generated across these five research questions is systematically fed into a GPT-4 API in our tool that users can then use to prompt with natural language questions about their text data.

This tool also lets users export their chat.db data into an easy-to-analyze CSV for custom workflows.

All analytical features described in this section are implemented in our open-source tool, available on GitHub (Gerber, 2025).

## 6 Limitations

The most significant limitation to further studies involving text message data is consent and privacy. Our tool aims to fix this by letting users select a de-identify participants option, that can let users safely share results and figures for any future studies. Another potential limitation is that of topic modeling, which produces topics in a non-human-readable format. This may be addressed using newer, alternate methods that rely on LLMs to identify topics and classify text (Mu et al., 2024).

## 7 Ethics Statement

All iMessage data analyzed in this study were provided voluntarily by the participants and processed entirely on their local devices. Identifying information was removed during preprocessing, and no data were shared externally. The anonymous participant's data were fully de-identified, and all analyses were conducted in accordance with standard ethical practices for handling personal communication data.